\documentclass[11pt,a4paper]{article}
\usepackage[hyperref]{acl2019}
\usepackage{times}
\usepackage{latexsym}
\usepackage{booktabs}
\usepackage{amsmath}
\usepackage[utf8]{inputenc}
\usepackage{url}

\usepackage{multirow}
\usepackage{rotating}
\aclfinalcopy % Uncomment this line for the final submission

\newenvironment{itemizesquish}{\begin{list}{\labelitemi}{\setlength{\topsep}{0.5em}\setlength{\itemsep}{0em}\setlength{\labelwidth}{0.5em}\setlength{\leftmargin}{\labelwidth}\addtolength{\leftmargin}{\labelsep}}}{\end{list}}

\title{Unbabel's Participation\\in the WMT19 Translation Quality Estimation Shared Task}

\author{F\'abio Kepler \\
  Unbabel 
 \And
  Jonay Tr\'enous \\
  Unbabel 
  \And
  Marcos Treviso \\
  Instituto de Telecomunica\c{c}\~oes
 \And 
  Miguel Vera \\
  Unbabel
  \AND
  António Góis \\
  Unbabel
\And
  M. Amin Farajian \\
  Unbabel
  \And
  António V. Lopes \\
  Unbabel
 \And
  Andr\'e F.~T.~Martins \\
  Unbabel 
\AND \\ 
  \texttt{\{kepler, sony, miguel.vera\}@unbabel.com}\\
  \texttt{\{antonio.gois, amin, antonio.lopes, andre.martins\}@unbabel.com}\\
  \texttt{marcosvtreviso@gmail.com}
}

\date{}

\begin{document}
\maketitle
\begin{abstract}
We present the contribution of the
Unbabel team to the WMT 2019 Shared
Task on Quality Estimation.
We participated on the word, 
sentence, and document-level tracks, encompassing 3 language pairs: English-German, English-Russian, and English-French.
Our submissions build upon the recent {\tt OpenKiwi} framework: we combine linear, neural, and predictor-estimator systems with new transfer learning approaches using BERT and XLM pre-trained models. We compare systems individually and propose new ensemble techniques for word and sentence-level predictions. We also propose a simple technique for converting word labels into document-level predictions. Overall, our submitted systems achieve the best results on all tracks and language pairs by a considerable margin. 
\end{abstract}

\section{Introduction}

Quality estimation (QE) is the task of evaluating a translation system's quality without access to reference translations~\cite{Blatz2004,Specia2018}. 
%Among its potential usages are:
%informing an end user about the reliability of automatically translated content;
%deciding if a translation is ready for publishing or if it requires human post-editing;
%and highlighting the words that need to be post-edited. 
This paper describes the contribution of the Unbabel team to the Shared Task on Word, Sentence, and Document-Level (QE Tasks 1 and 2) at WMT 2019. 

Our system 
adapts {\tt OpenKiwi},%
\footnote{\url{https://unbabel.github.io/OpenKiwi}.} %
a recently released open-source framework for QE that implements the best QE systems from WMT 2015-18 shared tasks  \cite{Martins2016,Martins2017TACL,Kim2017,Wang2018}, which we extend to leverage recently proposed pre-trained models via transfer learning techniques. Overall, our main contributions are as follows:
\begin{itemizesquish}
    \item We extend {\tt OpenKiwi} with a Transformer  predictor-estimator model \citep{Wang2018}.
    \item We apply transfer learning techniques, fine-tuning BERT \citep{devlin2018bert} and XLM \citep{lample2019cross} models in a predictor-estimator architecture.
    \item We incorporate predictions coming from the APE-BERT system described in \citet{apebert}, also used in this year's Unbabel's APE submission \citep{unbabelape}. 
    \item We propose new ensembling techniques for combining word-level and sentence-level predictions, which outperform previously used stacking approaches \citep{Martins2016}.
    \item We build upon our BERT-based predictor-estimator model to obtain document-level annotation and MQM predictions via a simple word-to-annotation conversion scheme.
\end{itemizesquish}

Our submitted systems achieve the best results on all tracks and all language pairs by a considerable margin: on English-Russian (En-Ru), our sentence-level system achieves a Pearson score of 59.23\% (+5.96\% than the second best system), and on English-German (En-De), we achieve 57.18\% (+2.44\%).

%\begin{figure*}[!t]
%\small
%\centering
%\includegraphics[width=\textwidth]{qe_example}
%\caption{\label{fig:qe}Example from the WMT 2018 word-level QE training set. Shown are the English source sentence (top), the German machine translated text (bottom), and its manual post-edition (middle). We show also the three types of word-level quality tags: MT (or target) tags account for words that are replaced or deleted, gap tags account for words that need to be inserted, and source tags indicate what are the source words that were omitted or mistranslated. For this example, the HTER sentence-level score (number of edit operations to produce PE from MT normalized by the length of PE) is $8/12 = 66.7\%$, corresponding to 4 insertions, 1 deletion, and 3 replacements out of 12 reference words.}
%\end{figure*}

% -------------------------------------------------------------------
\section{Word and Sentence-Level Task}\label{sec:models}
% -------------------------------------------------------------------

The goal of the  word-level QE task %(Figure~\ref{fig:qe}) 
is to assign quality labels (\textsc{ok} or \textsc{bad}) to each \emph{machine-translated word}, as well as to \emph{gaps} between words (to account for context that needs to be inserted), and \emph{source words} (to denote words in the original sentence that have been mistranslated or omitted in the target). 
%In the last years, the most accurate systems that have been developed for this task combine linear and neural models
%\cite{Kreutzer2015,Martins2016}, 
%use automatic post-editing as an intermediate step \citep{Martins2017TACL}, or develop specialized neural architectures \citep{Kim2017,Wang2018}. 
The goal of the 
Sentence-level QE task, on the other hand, is to predict the quality of the whole translated sentence, based on how many edit operations are required to fix it, in terms of HTER (Human Translation Error Rate) \citep{Specia2018}. 
We next describe the datasets, resources, and models that we used for these tasks.

%The most successful approaches to sentence-level QE to date are based on conversions from word-level predictions \citep{Martins2017TACL} or joint training with multi-task learning \citep{Kim2017,Wang2018}. 

\subsection{Datasets and Resources}

The data resources we use to train our systems are of three types: the QE shared task corpora, additional parallel corpora, and artificial triplets ({\tt src}, {\tt pe}, {\tt mt}) in the style of the eSCAPE corpus \cite{negri2018escape}.
\begin{itemizesquish}
    \item The En-De QE corpus provided by the shared task, consisting of 13,442 {\tt train}  triplets.
    \item The En-Ru QE corpus provided by the shared task, consisting of 15,089 {\tt train} triplets.
    \item The En-De parallel dataset of 3,396,364  sentences from the IT domain provided by the shared task organizers. %\footnote{Dataset can be found under Additional Resouces at \url{http://www.statmt.org/wmt19/qe-task.html}},  
    which we extend in the style of the eSCAPE corpus to contain artificial triplets. To do this, we use OpenNMT with 5-fold jackknifing \citep{Klein2017} to obtain unbiased translations of the source sentences.
    \item The En-Ru eSCAPE corpus \cite{negri2018escape} consisting of 7,735,361 artificial triplets.
\end{itemizesquish}
%\subsubsection{eSCAPE corpora}
%\label{sec:escape}
%To create an escape style corpus from the parallel English-German IT corpus, we perform the following steps:
%    \begin{enumerate}
%    \item Split the corpus into 5 folds $f_i$
%    \item Use OpenNMT \citep{Klein2017} to train 5 LSTM based translation models, one model $\mathcal{M}_i$ for every subset created by removing fold $f_i$ from the training data.
%    \item Translate each fold $f_i$ using the translation Model $\mathcal{M}_i$.
%    \item Join the translations to get an unbiased machine translated version of the full corpus.
%    \item Remove empty lines
%    \end{enumerate}

\subsection{Linear Sequential Model}\label{sec:linear}

Our simplest baseline is the linear sequential model described by \citet{Martins2016,Martins2017TACL}. It is a discriminative feature-based sequential model (called \textsc{LinearQE}).
The system 
%receives as input
%a tuple
%$\langle s, t, \mathcal{A}\rangle$,
%where %:
%$s = s_1 \ldots s_M$ is the source sentence,
%$t = t_1 \ldots t_N$ is the translated sentence,
%and
%$\mathcal{A} \subseteq \{(m,n)\,\,|\,\, 1 \le m \le M, \,\, 1 \le n \le N\}$
%is a set of word alignments.
%It predicts as output a sequence $\widehat{y} = y_1 \ldots y_N$, with each $y_i \in \{\textsc{bad}, \textsc{ok}\}$, by using a 
uses a first-order sequential model with unigram and bigram features, whose weights are learned by using the max-loss MIRA algorithm \cite{Crammer2006}. 
The features include information about the words, part-of-speech tags, and syntactic dependencies, obtained with  TurboParser \cite{Martins2013ACL}.

%----------------
\subsection{NuQE}

We used {\sc NuQE} (NeUral Quality Estimation) as implemented in {\tt OpenKiwi} \citep{openkiwiacl} and adapted it to jointly learn MT tags, source tags and also sentence scores.
We use the original architecture with the following additions.
For learning sentence scores, we first take the average of the MT tags output layer and than pass the result through a feed-forward layer that projects the result to a single unit. 
For jointly learning source tags, we take the source text embeddings, project them with a feed-forward layer, and then sum the MT tags output vectors that are aligned.
The result is then passed through a feed-forward layer, a bi-GRU, two other feed-forward layers, and finally an output layer.
The layer dimensions are the same as in the normal model. 
It is worth noting that {\sc NuQE} is trained from scratch using only the shared task data, with no pre-trained components, besides Polyglot embeddings \citep{al2013polyglot}.

\subsection{RNN-Based Predictor-Estimator}

Our implementation of the RNN-based prediction estimator ({\sc PredEst-RNN}) is described in \citet{openkiwiacl}. It follows closely the architecture proposed by \citet{Kim2017}, which consists of two modules:
\begin{itemizesquish}
    \item a \emph{predictor}, which is trained to predict each token of the target sentence given the source and the left and right context of the target sentence;
    \item an \emph{estimator}, which takes features produced by the \emph{predictor} and uses them to classify each word as {\sc ok} or {\sc bad}.
\end{itemizesquish} 
Our predictor uses a biLSTM to encode the source, and two unidirectional LSTMs processing the target in left-to-right (LSTM-L2R) and right-to-left (LSTM-R2L) order.
For each target token $t_i$, the representations of its left and right context are concatenated and used as query to an attention module before a final softmax layer.
It is trained on the large parallel corpora mentioned above.
The estimator takes as input a sequence of features: for each target token $t_i$, the final layer before the softmax (before processing $t_i$), and the concatenation of the $i$-th hidden state of LSTM-L2R and LSTM-R2L (after processing $t_i$).
We train this system with a multi-task architecture that allows us to predict sentence-level HTER scores.
Overall, this system is capable to predict sentence-level scores and all word-level labels (for MT words, gaps, and source words)---the source word labels are produced by training a predictor in the reverse direction.

\subsection{Transformer-Based Predictor-Estimator}

In addition, we implemented a Transformer-based predictor-estimator ({\sc PredEst-Trans}), following \citet{Wang2018}. This model has the following modifications in the \textit{predictor}: (i) in order to encode the source sentence, the bidirectional LSTM is replaced by a Transformer encoder; (ii) the LSTM-L2R is replaced by a Transformer decoder with future-masked positions, and the LSTM-R2L is replaced by a Transformer decoder with past-masked positions. Additionally, the Transformer-based model produces the ``mismatch features'' proposed by \citet{fan2018bilingual}.
% For the \textit{estimator} part, we tried replacing the BiLSTM layer by a linear layer or a 1D convolutional layer followed by max-pooling, but we kept the BiLSTM since it returned the best results.

\subsection{Transfer Learning and Fine-Tuning}
\label{sec:transfer}

Following the recent trend in the NLP community leveraging large-scale language model pre-training for a diverse set of downstream tasks, we used two pre-trained language models as feature extractors, the multilingual BERT \cite{devlin2018bert} and the Cross-lingual Language Model (XLM) \cite{lample2019cross}.  The predictor-estimator model consists of a predictor that produces contextual token representations, and an estimator that turns these representations into predictions for both word level tags, and sentence level scores. 
As both of these models produce contextual representations for each token in a pair of sentences, we simply replace the predictor part by either BERT or XLM to create new QE models: \textsc{PredEst-BERT} and \textsc{PredEst-XLM}. 
The XLM model is particularly well suited to the task at hand, as its pre-training objective already contains a translation language modeling part. 

For improved performance, we employ a pre-fine-tuning step by continuing their language model pre-training on data that is closer to the domain of the shared task. For the En-De pair we used the in-domain data provided by the shared task, and for the En-Ru pair we used the eSCAPE corpus \cite{negri2018escape}. 
% In order to use BERT and XLM without making modifications in our pipeline, we treat them as a \textit{predictor} model. This means that BERT and XLM were also finetuned for the QE tasks. 

Despite the shared multilingual vocabulary, BERT is originally a monolingual model, treating the input as either being from one language or another. We pass both sentences as input by concatenating them according to the template: \texttt{[CLS] target [SEP] source [SEP]}, where \texttt{[CLS]} and \texttt{[SEP]} are special symbols from BERT, denoting beginning of sentence and sentence separators, respectively. 
%We tried passing only one sentence at a time, but we got worse results. We also tried concatenating source before target, but the results were the same in average. 
In contrast, XLM is a multilingual model which receives two sentences from different languages as input. Thus, its usage is straightforward. 
 
 The output from BERT and XLM is split into target features and source features, which in turn are passed to the regular \textit{estimator}. They work with word pieces rather than tokens, so the model maps their output to tokens by selecting the first word piece of each token. For En-Ru the mapping is slightly different, it is done by taking the average of the word pieces of each token.
%  , and the mapping was done after we calculated the logits.
 
 For \textsc{PredEst-BERT}, we obtained the best results by ignoring features from the other language, that is, for predicting target and gap tags we ignored source features, and for predicting source tags we ignored target features. On the other hand, \textsc{PredEst-XLM} predicts labels for target, gaps and source at the same time. As the predictor-estimator model, \textsc{PredEst-BERT} and \textsc{PredEst-XLM} are trained in a multi-task fashion, predicting sentence-level scores along with word-level labels.

% BERT-based models were implemented by using the pretrained weights from \cite{devlin2018bert}.

\subsection{APE-QE}

In addition to traditional QE systems, we also use Automatic Post-Editing (APE) adapted for QE (\textsc{APE-QE}), following \citet{Martins2017TACL}. An APE system is trained on the human post-edits and its outputs are used as pseudo-post-editions to generate word-level quality labels and sentence-level scores in the same way that the original labels were created.

We use two variants of {\sc APE-QE}: 
\begin{itemizesquish}
\item {\sc Pseudo-APE}, which trains a regular translation model and uses its output as a pseudo-reference.
\item An adaptation of BERT to APE ({\sc APE-BERT}) with an additional decoding constraint to reward or discourage words that do not exist in the source or MT.
\end{itemizesquish}

{\sc Pseudo-APE} was trained using {\tt OpenNMT-py} \citep{Klein2017}. For En-De, we used the IT domain corpus provided by the shared task, and for En-Ru we used the Russian eSCAPE corpus \citep{negri2018escape}. 

For {\sc APE-BERT}, we follow the approach of~\newcite{apebert}, also used by Unbabel's APE shared task system~\citep{unbabelape}, and adapt BERT to the APE task using the QE in-domain corpus and the shared task data as input, where the source and MT sentences are the encoder's input and the post-edited sentence is the decoder's output. 
In addition, we also employ a conservativeness penalty~\citep{unbabelape}, a beam decoding penalty which either rewards or penalizes choosing tokens not in the \texttt{src} and \texttt{mt}, with a negative score to encourage more edits of the MT.

\subsection{System Ensembling} 

We ensembled the systems above to produce a single prediction, as described next.

\paragraph{Word-level ensembling.}
We compare two approaches:
\begin{itemizesquish}
    \item A stacked architecture with a feature-based linear system, as described by \citet{Martins2017TACL}. This approach uses the predictions of various systems as additional features in the linear system described in \S\ref{sec:linear}. To avoid overfitting on the training data, this approach requires jackknifing. 
    \item A novel strategy consisting of learning a convex combination of system predictions, with the weights learned on the development set. We use Powell's conjugate direction method \citep{powell1964}%
    \footnote{This is the method underlying the popular MERT method  \citep{mert2003}, widely used in the MT literature.} %
    as implemented in {\tt SciPy} \citep{scipy2001} to directly optimize for the task metric ($F_1$-MULT).
\end{itemizesquish}
Using the development set for learning carries a risk of overfitting; by using $k$-fold cross-validation we avoided this, and indeed the performance is equal or superior to the linear stacking ensemble (\autoref{tab:ensemble}), while being computationally cheaper 
as only the development set is needed to learn an ensemble, avoiding jackknifing. 
% (linear stacking requires training each ensembled system in a jackknifed fashion to produce unbiased predictions on the full training set).

\begin{table}[t]
    \centering
    \begin{tabular}{l c} \toprule
        \textsc{Method} & \textsc{Target} $F_1$ \\ \midrule
         \textsc{Stacked Linear} & 43.88  \\
         \textsc{Powell} & 44.61 \\ \bottomrule
    \end{tabular}
    \caption{Performance of the stacked linear ensemble and  Powell's method on the WMT17 dev set ($F_1$-MULT on MT tags). The ensemble is over the same set of models\footnote{available via \url{https://unbabel.github.io/OpenKiwi/reproduce.html}} reported in the release of the OpenKiwi \citep{openkiwiacl} framework. To estimate the performance of Powell's method, the dev set was partitioned into 10 folds $f_i$. We ran Powell's method 10 times, leaving out one fold at a time, to learn weights $w_i$. Predicting on fold $f_i$ using weights $w_i$ and calculating $F_1$ performance over the concatenation of these predictions gives an approximately unbiased estimate of the performance of the method.}
    \label{tab:ensemble}
\end{table}

\paragraph{Sentence-level ensembling.}
We have systems outputting sentence-level predictions directly, and others outputting word-level probabilities that can be turned into sentence-level predictions by averaging them over a sentence, as in \citep{Martins2017TACL}. To use all available features (sentence score, gap tag, MT tag and source tag predictions from all systems used in the word-level ensembles), we learn a linear combination of these features using $\ell_2$-regularized regression over the development set. We tune the regularization constant 
%we first run an exponential search to find a rough estimate of a good value, then a linear search in a multiplicative neighborhood of the best value found in the first round. As the development set is our training set, we again 
with $k$-fold cross-validation, %to estimate the performance of the algorithm with a given regularization constant, and finally 
and retrain on the full development set using the chosen value. 

\section{Document-Level Task}

Estimating the quality of an entire document introduces additional challenges. 
The text may become too long to be processed at once by previously described methods, and longer-range dependencies may appear (e.g inconsistencies across sentences).

Both sub-tasks were addressed: estimating the MQM score of a document and identifying character-level annotations with corresponding severities. Note that, given the correct number of annotations in a document and their severities, the MQM score can be computed in closed form. However, preliminary experiments using the predicted annotations to compute MQM did not outperform the baseline, hence we opted for using independent systems for each of these sub-tasks.

\subsection{Dataset}

The data for this task consists of Amazon reviews translated from English to French using a neural MT system. Translations were manually annotated for errors, with each annotation associated to a severity tag (minor, major or critical). %and a category (out of 43 possible ones). Predicted annotations on the test set must include a severity prediction but not a category prediction, although the category may provide additional information when training a Quality Estimation system.

\begin{figure}[t]
\small
\centering
\includegraphics[width=\columnwidth]{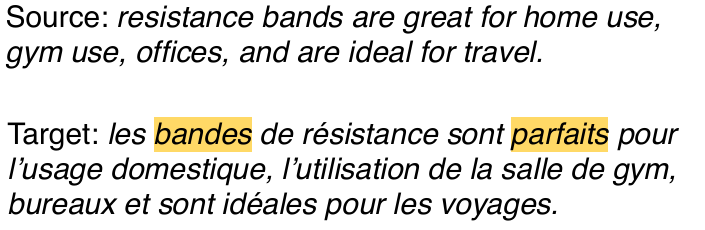}
\caption{\label{fig:ann}Example of a multi-span annotation containing two spans: \textit{parfaits} does not agree with \textit{bandes} due to gender---it should be \textit{parfaites}. This mistake corresponds to a single annotation with severity ``minor".}
\label{fig:annotation}
\end{figure}

Note that each annotation may include several words, which do not have to be contiguous. We refer to each contiguous block of characters in an annotation as a span, and refer to an annotation with at least two spans as a multi-span annotation. \autoref{fig:annotation} illustrates this, where a single annotation is comprised of the spans \textit{bandes} and \textit{parfaits}.

Across training set and last year's development and test set, there are 36,242 annotations. Out of these, 4,170 are multi-span, and 149 of the multi-span annotations contain spans in different sentences. The distribution of severities is 84.12\% of major, 11.74\% of minor and 4.14\% of critical.

\subsection{Implemented System}

To predict annotations within a document the problem is first treated as a word-level task, with each sentence processed separately. To obtain gold labels, the training set is tokenized and an {\sc ok}/{\sc bad} tag is attributed to each token, depending on whether the token contains characters belonging to an annotation. Note that besides token tags, we will also have gap tags in between tokens. A gap tag will only be labeled as {\sc bad} if a span begins and ends exactly in the borders of the gap. Our best-performing model for the word-level part is an ensemble of 5 BERT models. Each BERT model was trained as described in \S\ref{sec:transfer}, but without pre-fine-tuning. Systems were ensembled by a simple average.

Later, annotations may be retrieved from the predicted word-level tags by concatenating contiguous {\sc bad} tokens into a single annotation. This is done for token-tags, while each gap-tag can be directly mapped to a single annotation without attempting any merge operation. Note that this immediately causes 4 types of information loss, which can be addressed in a second step:
\begin{itemizesquish}
    \item Severity information is lost, since all three severity labels are converted to {\sc bad} tags. As a baseline, all spans are assigned the most frequent severity, ``major."
    \item Span borders are defined on character-level, whose positions may not match exactly the beginning or ending of a token. This will cause all characters of a partially correct token to be annotated with an error.
    \item Contiguous {\sc bad} tokens will always be mapped to a single annotation, even if they belong to different ones.
    \item Non-contiguous {\sc bad} tokens will always be mapped to separate annotations, even if they belong to the same one.
\end{itemizesquish}
Although more sophisticated approaches were tested for predicting severities and merging spans into the same annotation, these approaches did not result in significant gains, hence we opted by using the previously described pipeline as our final system. 
%Our final system for predicting annotations uses the previously described pipeline, . To predict either severities of spans or whether pairs of spans should be merged into a single annotation, a span representation was devised, inspired by \citet{wang2016graph}. We first obtain token-level embeddings from a BERT model fine-tuned for the quality estimation task. Given span's boundaries, the tokens of a sentence are split into 3 groups: left context, span, and right context. The representation of each of the 3 is obtained, either by subtracting the first token to the last one or by averaging all tokens. After concatenating the 3 embeddings we have a span representation, which is used together with a binary variable (is gap) to classify spans: both their severities and whether each pair should be connected. However these approaches degrade results on F1 annotations, compared to baselines of never merging spans and always predicting 'major' as the severity. This can be explained by the rarity of the other categories and of multi-span annotations in the data. Also, the gains of using the correct merges and severities were measured experimentally, and only provided residual gains. 
To predict document-level MQM, each sentence's MQM is first predicted and used to get the average sentence MQM (weighting the average by sentence length degraded results in all experiments). This is used together with 3 percentages of BAD tags from the word-level model (considering token tags, gap tags and all tags) as features for a linear regression which outputs the final document-level MQM prediction. The percentage of BAD tags is obtained from the previously described word-level predictions, whereas the sentence MQMs are obtained from an ensemble of 5 BERT models trained for sentence-level MQM prediction. Again, each BERT model was trained as described in \S\ref{sec:transfer} without pre-fine-tuning, and the ensembling consisted of a simple average.%
\footnote{Using the approach of \citet{ive2018sheffield} proved less robust to this year's data due to differences in the annotations. Particularly some outliers containing zero annotations would strongly harm the final Pearson score when mis-predicted.} %

\begin{table}[t]
\begin{center}
\scriptsize
\begin{tabular}{llccc}
\toprule
\textsc{Pair} & \textsc{System} & \textsc{Target} $F_1$ & \textsc{Source} $F_1$ & \textsc{Pearson} \\
\midrule
\multirow{8}{*}{En-De}  & \textsc{Linear} & 0.3346 & 0.2975 & - \\
                        & APE-QE 		&   0.3740   &   0.3446   & 0.3558  \\ 
                        & APE-BERT 		&   0.4244   &   0.4109   & 0.3816  \\ 
                        & \textsc{PredEst-RNN}      &   0.3786  &   -        & 0.5020      \\
                        & \textsc{PredEst-Trans}	&   0.3980   &   -        & 0.5300  \\ 
                        & \textsc{PredEst-XLM} 		&   0.4144   &   0.3960   & 0.5810  \\ 
                        & \textsc{PredEst-BERT} 	&   0.3870   &   0.3310   & 0.5190  \\ 
                        & \textsc{Linear Ens.} 		&   0.4520   &   0.4116   & -  \\  
                        & (*)\textsc{Powell's Ens.} 	&   0.4872   &   0.4607   & 0.5968  \\

\midrule
\multirow{9}{*}{En-Ru}  & \textsc{Linear} & 0.2839 & 0.2466 & - \\
                        & APE-QE 		& 	0.2592   &   0.2336   & 0.1921  \\
                        & APE-BERT 		& 	0.2519   &   0.2283   & 0.1814  \\
                        & \textsc{NuQE} & 	0.3130   &   0.2000   & -       \\
                        & \textsc{PredEst-RNN}   &   0.3201   &   -        & -       \\
                        & \textsc{PredEst-Trans} & 	0.3414   &   -        & 0.3655  \\
                        & \textsc{PredEst-XLM} 	 & 	0.3799   &   0.3280   & 0.4983  \\
                        & \textsc{PredEst-BERT}  & 	0.3782   &   0.3039   & 0.5000  \\
                        & (*)\textsc{Ensemble 1} 	&   0.3932   &   0.3640   & 0.5469   \\
                        & (*)\textsc{Ensemble 2} 	&   0.3972   &   0.3700   & 0.5423    \\

\bottomrule
\end{tabular}
\end{center}
\caption{Word and sentence-level results for En-De and En-Ru on the validation set in terms of $F_1$-MULT and Pearson's $r$ correlation. (*) Lines with an asterisk  use Powell's method for word level ensembling and $\ell_2$-regularized regression for sentence level. As the weights are tuned on the dev set, their numbers can not be directly compared to the other models} \label{tab:wl_sl_results_dev}
\end{table}	

\begin{table*}[!htb]
\small
\begin{center}
\begin{tabular}{llccccc}
\toprule
\textsc{Pair} & \textsc{System} & \textsc{Target} $F_1$ & \textsc{Target} $MCC$ & \textsc{Source} $F_1$ & \textsc{Source} $MCC$ & \textsc{Pearson} \\
\midrule

\multirow{3}{*}{En-Ru}   & \textit{Baseline}     & 0.2412   & 0.2145 &   0.2647  & 0.1887 &  0.2601 	\\
                         & \textsc{Ensemble 1} 	 & 0.4629   & 0.4412 &   0.4174  & 0.3729 &  0.5889  	\\
                         & \textsc{Ensemble 2} 	 & 0.4780   & 0.4577 &   0.4541  & 0.4212 &  0.5923  	\\

\midrule

\multirow{3}{*}{En-De}   & \textit{Baseline}             & 0.2974 & 0.2541 & 0.2908 & 0.2126 & 0.4001 \\
                         & \textsc{Linear Ensemble} 	 & 0.4621 & 0.4387 & 0.4284 & 0.3846 & - \\
                         & \textsc{Powell's Ensemble} 	 & 0.4752 & 0.4585 & 0.4455 & 0.4094 & 0.5718 \\

\bottomrule
\end{tabular}
\end{center}
\caption{Word and sentence-level results for En-De and En-Ru on the test set in terms of $F_1$-MULT and Pearson's $r$ correlation.} \label{tab:wl_sl_results_test}
\end{table*}

\section{Experimental Results}

\subsection{Word and Sentence-Level Task}

The results achieved by each of the systems described in \S\ref{sec:models} for En-De and En-Ru on the validation set are shown in Table~\ref{tab:wl_sl_results_dev}. 
We tried the following strategies for ensembling:
\begin{itemizesquish}
\item For En-De, we created a word-level ensembled system with Powell's method, by combining one instance of the {\sc APE-BERT} system, another instance of the {\sc Pseudo-APE-QE} system, 10 runs of the {\sc PredEst-XLM} model (trained jointly for all subtasks), 6 runs of the same model without pre-fine-tuning, 5 runs of the {\sc PredEst-BERT} model (trained jointly for all subtasks), and 5 runs of the  {\sc PredEst-Trans} model (trained jointly for MT and sentence subtasks, but not for predicting source tags). For comparison, we report also the performance of a stacked linear ensembled word-level system. 
For the sentence-level ensemble, we learned system weights by fitting a linear regressor to the sentence scores produced by all the above models.
\item For En-Ru, we tried two versions of word-level ensembled systems, both using Powell's method: {\sc emsemble 1} combined one instance of the {\sc APE-BERT} system, 5 runs of the {\sc PredEst-XLM} model (trained jointly for all subtasks), one instance of the {\sc PredEst-BERT} model (trained jointly for all subtasks), 5 runs of the {\sc NuQE} models (trained jointly for all subtasks), and 5 runs of the  {\sc PredEst-Trans} model (trained jointly for MT and sentence subtasks, but not for predicting source tags). {\sc emsemble 2} adds to the above predictions from the {\sc Pseudo-APE-QE} system. In both cases, for sentence-level ensembles, we learned system weights by fitting a linear regressor to the sentence scores produced by all the above models.
\end{itemizesquish}

The results in Table~\ref{tab:wl_sl_results_dev} show that the transfer learning approach with BERT and XLM benefits the QE task. The \textsc{PredEst-XLM} model, which has been pre-trained with a translation objective, has a small but consistent advantage over both \textsc{PredEst-BERT} and \textsc{PredEst-Transf}. A clear takeaway is that ensembling of different systems can give large gains, even if some of the subsystems are weak individually.

% What can we see from the table?
% =============
% For EN-RU:
% ----
% XLM and BERT have high scores for all metrics
% NuQE, \textsc{PredEst-Trans} and APE have lower results, suggesting that the transfer-learning approach was fundamental
% By using BPEs to train the predictor part from \textsc{PredEst-Trans}, we didn't observe major improvements, showing that the superiority of BERT and XLM are not related to the wordpieces strategy. 
% Big gain by ensembling them, in average 2 points for target, 6-7 for source, and 4-5 points for pearson.

% For EN-DE
% ----
% \textsc{PredEst-Trans} was able to achieve good results due to the similarity of indomain corpus with the shared task corpus. The use of mismatch features gave us a boost of 1 points for word level, 2 points for sentence level scores.
% XLM alone was able to get very high results. By tuning parameters for sentence level, it was able to achieve results close to the powell's ensemble.
% Finally, as for EN-DE, by ensembling these models we were able to achieve very high results, increasing 6-7 points for word-level scores.

Table~\ref{tab:wl_sl_results_test} shows the results obtained with our ensemble systems on the official test set.

\subsection{Document-Level Task}

Finally, \autoref{tab:doc_results} contains results for document-level submissions, both on validation and test set submissions. On $F_1$ annotations, results across all data sets are reasonably consistent. On the other hand, MQM Pearson varies significantly between dev and dev0. 
%Pearson seems to be particularly sensitive to outliers that have zero annotations, 
Differences in the training of the two systems shouldn't explain this variation, since both have equivalent performance on the test set.

\begin{table}[t]
\begin{center}
\small
\begin{tabular}{lccc}
\toprule
               & \textsc{Dev}    & \textsc{Dev0}   & \textsc{Test}   \\
\midrule
$F_1$ \textsc{ann.} (BERT) & 0.4664 & 0.4457 & 0.4811 \\
MQM (BERT)     & 0.3924 & -      & 0.3727 \\
MQM (LINBERT)  & -      & 0.4714 & 0.3744 \\
\bottomrule
\end{tabular}
\caption{Results of document-level submissions, and their performance of the dev and dev0 validation sets. %Making use of both development sets available (dev and dev0), the MQM linear regression was trained on each of them - BERT submission was trained on dev0, LINBERT submission on dev
}
\label{tab:doc_results}
\end{center}
\end{table}

% -------------------------------------------------------------------
\section{Conclusions}
% -------------------------------------------------------------------

We presented Unbabel's contribution to the WMT 2019 Shared
Task on Quality Estimation. 
Our submissions are based on the {\tt OpenKiwi} framework, to which we added new transfer learning approaches via BERT and XLM pre-trained models. We also proposed a new ensemble technique using Powell's method that outperforms previous strategies, and we convert word labels into span annotations to obtain document-level predictions. Our submitted systems achieve the best results on all tracks and language pairs. 

% -------------------------------------------------------------------
\section*{Acknowledgments}
% -------------------------------------------------------------------

The authors would like to thank  the  support  provided by the EU in the context of the PT2020 project (contracts 027767 and 038510), by  the  European  Research  Council  (ERC  StG  DeepSPIN  758969), and  by  the  Fundação  para  a  Ciência  e  Tecnologia through contract UID/EEA/50008/2019.

% -------------------------------------------------------------------
\bibliography{acl2019}
\bibliographystyle{acl_natbib}

\end{document}